\documentclass{article}

\PassOptionsToPackage{numbers, compress}{natbib}

\usepackage[preprint]{neurips_2022}

\usepackage[utf8]{inputenc} 
\usepackage[T1]{fontenc}    
\usepackage{hyperref}       
\usepackage{url}            
\usepackage{booktabs}       
\usepackage{amsfonts}       
\usepackage{nicefrac}       
\usepackage{microtype}      
\usepackage{xcolor}         
\usepackage{natbib}
\usepackage{multirow}
\usepackage{pgfplots}
\usepackage{amsmath}
\usepackage{subcaption}
\pgfplotsset{compat=1.18}

\usepackage{graphics} 
\title{Deep evolving semi-supervised anomaly detection}

\author{%
  Jack Belham \\
  Oxford-Brookes University \\
  Oxford, UK \\
  \And
  Aryan Bhosale \\
  Indian Institute of Technology Bombay \\
  Mumbai, India \\
  \And
  Samrat Mukherjee \\
  Indian Institute of Technology Bombay \\
  Mumbai, India \\
  \And
  Biplab Banerjee \\
  Indian Institute of Technology Bombay \\
  Mumbai, India \\
  \And
  Fabio Cuzzolin \\
  Oxford-Brookes University \\
  Oxford, UK \\
}

\begin{document}

\maketitle

\begin{abstract}
The aim of this paper is to formalise the task of continual semi-supervised anomaly detection (CSAD), with the aim of highlighting the importance of such a problem formulation which assumes as close to real-world conditions as possible. After an overview of the relevant definitions of continual semi-supervised learning, its components, anomaly detection extension, and the training protocols; the paper introduces a baseline model of a variational autoencoder (VAE) to work with semi-supervised data along with a continual learning method of deep generative replay with outlier rejection. The results show that such a use of extreme value theory (EVT) applied to anomaly detection can provide promising results even in comparison to an upper baseline of joint training. The results explore the effects of how much labelled and unlabelled data is present, of which class, and where it is located in the data stream. Outlier rejection shows promising initial results where it often surpasses a baseline method of Elastic Weight Consolidation (EWC). A baseline for CSAD is put forward along with the specific dataset setups used for reproducability and testability for other practitioners. Future research directions include other CSAD settings and further research into efficient continual hyperparameter tuning.
\end{abstract}

\section{Introduction}

A long-standing task within Machine Learning (ML) has been that of anomaly detection (AD), defined as "the problem of finding patterns in data that do not conform to expected behaviour" \cite{anomaly_def}. Detecting anomalies within a system is a very important problem to solve within multiple different areas, such as fault detection in manufacturing \cite{xiao2016robust}, and healthcare \cite{cabral2014one}. Representative data collection is a particularly difficult task, where finding positive anomalous data may prove difficult or impossible depending on the dataset \cite{continual_learning_def}. This leads to another issue where these anomalies may change over time as the distribution of the data shifts. One way to mitigate this is to use continual learning \cite{continual_learning_def} to account for a dynamically shifting data distribution over time. This allows for new types of anomalies to be identified as the model is continually learning from a constant data stream.

Typically, unsupervised learning is used for anomaly detection. However, in a real-world setting, there are quite often samples that can be labelled by an expert, even if at a high cost. Utilising these labelled samples can improve model performance as seen in experiments run by Ruff \textit{et al}. \cite{ruff2019deep}. Applying a semi-supervised \cite{chapelle2009semi} approach which leverages the small amount of labelled data available with a continual learning approach to an anomaly detection task enables the model to adapt to a continuously shifting data distribution. 

Continual semi-supervised learning is a new learning paradigm initially formalised by Shahbaz \textit{et al}. \cite{fabiocssl} in 2021. This explores the overlap between continual learning \cite{continual_learning_def} and semi-supervised learning \cite{chapelle2009semi}. Within a continual learning setting, instead of one dataset, there are instead a number of disjoint datasets where a model only has access to the latest dataset at a given point in time, and cannot access previously seen datasets \cite{continual_learning_def}\cite{liu2020learning}. This creates a paradigm in which the model aims to mitigate catastrophic forgetting \cite{french1999catastrophic} in which a model forgets about a previous data distribution or "task" that it has previously learned. There have been numerous different techniques proposed to mitigate catastrophic forgetting within this setting \cite{ewc_cl},\cite{autoencoder_continual_anomaly_detection}.

The typical continual learning setting is that of supervised continual learning, where all of the data is labelled \cite{bagus2022beyond}. Within this setting, continual learning is a very well-researched area as demonstrated by the numerous libraries designed specifically for this task \cite{lomonaco2021avalanche}, \cite{douillard2021continuum}, \cite{normandin2021sequoia}. However, when this assumption is tightened and there is instead a mixture of labelled and unlabelled data within these disjoint datasets, the problem becomes CSSL \cite{fabiocssl}, a much less explored area of research. This setup provides a much more realistic setting for anomaly detection where anomalies are often not labelled due to their very nature of not being known. Additionally, it is often infeasible to store all of the training data for a model at one time due to storage limitations \cite{hashem2015rise}, thereby making a continual learning method very favourable within this situation. 

Continual Semi-supervised Anomaly detection is therefore being formulated within this paper as the overlap between CSSL and AD - combining together their inherent difficulties together for a close to real-world setting of anomaly detection.

\section{Related Works}

\subsection{Semi-supervised Learning}

In semi-supervised learning, a small proportion of labelled data is available alongside a large proportion of unlabelled data. Given a dataset \(X\) := \((x_i)\)  \(\forall i \in [1,..., n]\), we define \(X_l:= (x_1,...,x_l)\), where \(l<<n\)  as the labelled dataset and \(Y_l:= (y_1,...,y_l)\) as the corresponding labels. The unlabelled dataset is then defined as \(X_u:= (x_l+1,...,x_l+u)\),where \(l+u = n\). \cite{chapelle2009semi}

This technique leverages the labelled data samples to improve the performance of the model beyond that of an unsupervised model. Some examples of experiments run for deep SSAD can be seen in Ruff \textit{et al}. \cite{ruff2019deep}, confirming that results on MNIST and CIFAR-10 are improved through using labels when present.

According to Ouali \textit{et al}. \cite{ouali2020overview}, there are three main assumptions of semi-supervised learning. The \textbf{Smoothness assumption} states that if inputs within a high density zone are close, so should their output. The opposite holds also. The \textbf{Cluster assumption} states that if inputs \(x_1, x_2\) are in the same cluster, they are likely to be in the same class. Finally, the \textbf{Manifold assumption} states that high dimensional data lie roughly on a low-dimensional manifold.

\textbf{Consistency Regularisation} \\
Consistency regularisation enforces the cluster assumption in the model. All of the corresponding techniques rely upon the fact that realistic data augmentation applied to unlabelled data should not change the prediction of the model \cite{ouali2020overview}. The aim is to minimise the distance between two outputs \(f(x_1), f(x_2) \forall x_1 \in X_u\) and \(x_2\) a perturbed version of \(x_1\). Common distance metrics used are mean square error (MSE) and Kullback-Leibler Divergence (KL) \cite{kullback1997information}.

\textbf{Proxy-label methods} \\
According to Ouali \textit{et al}. \cite{ouali2020overview}, proxy label methods use a prediction model to create pseudo-labels for unlabelled data. The techniques used vary by how the pseudo-label is created. In self-training, the model produces the pseudo-labels itself. This is in opposition to multi-view learning where the labels are created by models which are trained on different views of the data. 

\textbf{Generative Models} \\
Generative models instead try to estimate the joint distribution across the dataset, including the corresponding labels. In the case of a variational autoencoder, neural networks are used to approximate the distribution of the data. It is vital that all parts of the model are smooth functions such that gradient descent can be performed on the model in order to train a variational autoencoder \cite{kingma2019introduction}. To adapt a variational autoencoder, which is typically trained in a supervised or unsupervised manner, an additional classifier is added to the network to optionally classify inputs where the class is missing \cite{kingma2019introduction}. 

Another popular generative method is a general adversarial network. Originally put forward to be trained in an unsupervised manner \cite{goodfellow2020generative}, it consists of a generator and a discriminator where the the goal of each is to generate data to fool the discriminator and to discriminte real from generated data respectively. The resulting model, after adaptation to enable a semi-supervised setting \cite{odena2016semi}, is capable of modelling the joint distribution of the dataset with remarkable accuracy. 

\subsection{Continual Learning}

According to Liu \cite{continual_learning_def}, the goal of continual learning is to learn a sequence of tasks without access to past data. A model is created with the initial data and tasks. The learner then receives a sequence of experiences which contains a subset of the overall data distribution and tasks. The goal is then to incrementally update the model exploiting information from a time series of unlabelled data points. The target domain itself may change over time, for example in discrete asynchronous steps, e.g. if a new building is constructed in the field of view of a camera \cite{fabiocssl}. 

\textbf{Regularisation Methods}

Elastic Weight Consolidation (EWC) was first developed by Kirkpatrick \textit{et al}. \cite{ewc_cl}. This method selectively constrains the model weights which are most important to a specific task. In the domain-incremental setting, we want a trade-off between plasticity and stability such that the model adapts to the changing data distribution, but does not do so too quickly so as to forget the previously seen data distribution. This is implemented as a baseline method in which to compare other continual learning methods against. 

\textbf{Replay Methods}

In replay methods, previously seen examples are stored according within a replay buffer, often limited through size constraints. The methods vary in how they decide which samples to choose to replay in later experiences; a similar choice of which data to label is made in active learning. Some such examples include the CLEAR method \cite{rolnick2019experience} which leverages off policy learning behavioural cloning to enhance stability in the stability-plasticity trade-off. One of the main disadvantages of replay methods is that it is often infeasible in real-world conditions to continuously add to a replay buffer due to potential storage limitations of the buffer itself.

\textbf{Generative Methods}

Similarly to how a VAE can be used for semi-supervised learning, it is also an effective generative replay method in continual learning \cite{wiewel2020continual},\cite{egorov2021boovae}\cite{shin2017continual}. It is possible to conditionally generate data using labels with a VAE (e.g. using MNIST and being able to generate a given digit). However, there are significant drawbacks to this approach: due to the stochastic nature of a VAE, one cannot control the quality of the samples being generated. This is an issue as data is sampled that is not representative of a given class and therefore over time, the class means of the latent space will drift from their original class centres. This will result in lower quality images being replayed over time and eventually lead to lower accuracy and incorrect modelling of the overall posterior. There are several options to mitigate this problem using exemplar rehearsal techniques. Examples of this include the use of core sets \cite{bachem2015coresets}, uniform sampling \cite{deja2021binplay} and nearest mean classifiers \cite{mensink2012metric}. However, a more elegant solution was recently put forward by Mundt \textit{et al}.  \cite{mundt2022unified}, who proposed calculating the inlier and outlier probability of a given point in the latent space and then sampling with rejection.

\subsection{Anomaly Detection}

\textbf{Semi Supervised Anomaly Detection} \\
According to Villa-Perez \textit{et al}. \cite{villa2021semi}, there are 29 state-of-the-art SSAD algorithms for anomaly detection. Amongst these are methods based on K Nearest Neighbours (KNN), GANs, VAEs, isolation forests and ensemble based methods. The state of the art is currently a Bagging-Random Miner when the algorithms were tested across 95 different datasets and the average AUC taken. This method is an ensemble based method which is domain-specific in masquerade detection. \cite{camina2019bagging} These algorithms include one class SVMS (ocSVM), isolation forests, and KNNs. Other deep generative methods are mentioned in other papers which analyse the effectiveness of semi-supervised anomaly detection such as DeepSAD \cite{villa2021semi}. 

\textbf{Anomaly Detection in Continual Learning} \\
There are many different examples of anomaly detection in a continual learning setting, indicating its importance as a developing area of research \cite{stocco2020towards},\cite{frikha2021arcade},\cite{hemati2021continual}. VAEs are used along with EWC and Experience replay as CL methods in \cite{hemati2021continual} as a means of continual anomaly detection. A simple VAE used with MLP in the encoder and decoder layers are effective due to the tabular nature of the data being used. Frikha \textit{et al.} (2020) use meta learning to approximate a continual anomaly detection model in their efficient implementation of continual learning. Stocco \textit{et al.} (2020) create an entire framework utilising a main system which does online continual anomaly detection, along with a drift detector and monitor which work out which samples are required to store within a buffer. Another method used within continual learning is deep generative replay using a VAE trained on unsupervised normal data \cite{autoencoder_continual_anomaly_detection}. Since the model was only trained on normal data,  the reconstruction probability of anomalous samples is then high, making it a good metric to use within generative models with reconstruction components.

\subsection{Continual Semi-Supervised Learning}

Continual semi-supervised learning (CSSL) was first formalised by Shahbaz \textit{et al}. \cite{fabiocssl}. In their definition of CSSL, an initial labelled training batch is available to first train a model before it is then incrementally updated from an unlabelled data stream. In this instance, one is closer to the real-world instance where there is less labelled data available, and one cannot assume that the labels are all correct. The difficult problems to be solved in semi-supervised learning and continual learning are compounded together into CSSL. Some such problems that need to be solved are those of catastrophic forgetting, plasticity-stability trade-off, class imbalance, especially in the instance of anomaly detection, and label noise. 

However, in an alternative definition of CSSL \cite{wang2021ordisco}, there are a mixture of labelled and unlabelled training data in each experience. This is another option for the formulation of the CSSL problem definition and is one that will be adopted for the rest of this paper. The reason for this is that the formulation in \cite{fabiocssl} is quite a unique setting for CSSL, whereas the generalised setting in \cite{wang2021ordisco} can be applied to more situations. 

\subsection{Techniques}

Due to the recent formulation of CSSL, there are few existing research papers within this domain, but the existing techniques will be covered here. As a baseline method put forward by  \cite{fabiocssl}, self training uses the model to create pseudo-labels which, if confident enough, are accepted as ground truth labels to enable the model to train in a supervised way - thereby reducing the problem to a continual supervised learning problem. Alternatively, conditional Triple-GANs were used by \cite{wang2021ordisco} in order to accept optionally labelled data using a classifier, generator and discriminator to model the joint distribution of the data for use in generative replay. This proved incredibly effective, albeit with few comparisons beside memory buffers which are often seen as relaxing the constraints of continual learning, rather than directly solving the problem. 

\section{Methodology}
\label{Chap3}

The aim of this paper is to define CSAD as a research problem which can be formulated as follows. 

Given a dataset \(D\) := \((d_i) \forall i \in [1,..., n]\), where each \(d_i\) is a disjoint dataset, or experience, whose union equals the original dataset, \(D\). A given experience is defined as follows: 

\(X\) := \((x_j)\) \(\forall j \in [1,..., k]\), we define \(X_l:= (x_1,...,x_l)\), where \(l<<k\)  as the labelled dataset and \(Y_l:= (y_1,...,y_l)\) as the corresponding labels, where \((y_i) \in \{0,1\}  \forall i \in [1,...,l]\). The unlabelled dataset is then defined as \(X_u:= (x_l+1,...,x_l+u)\), where \(l+u = k\). 

This sequence of experiences is then passed to a learner in order to correctly classify the labelled and unlabelled data as anomalous or normal, where only the most recent experience is available during training and no access to past data is allowed. The aim is to mitigate catastrophic forgetting in a set-up similar to continual semi-supervised domain incremental learning \cite{fabiocssl}. 

Whilst domain-incremental continual learning, semi-supervised learning and anomaly detection have all been researched as separate research areas \cite{fabiocssl},\cite{zhu2005semi},\cite{continual_learning_def}, no paper has yet defined the problem of continual semi-supervised anomaly detection. This is a very relevant problem as it approaches real-world conditions for anomaly detection without ignoring any labelled data that a learner might be able to use as in unsupervised learning. Additionally, with the rise of Big Data, there is often not enough space to store all of the past data in order to train on it at once \cite{hashem2015rise}. Additionally, it is very computationally inefficient to refresh a model at regular intervals as opposed to continually training on the data as a stream \cite{computational_inefficiencies}. All of these reasons point to the necessity for CSAD within a research and real-world setting. 

\subsection{Semi-supervised Variational Autoencoder}

\begin{figure}[t]
\includegraphics[width=14cm]{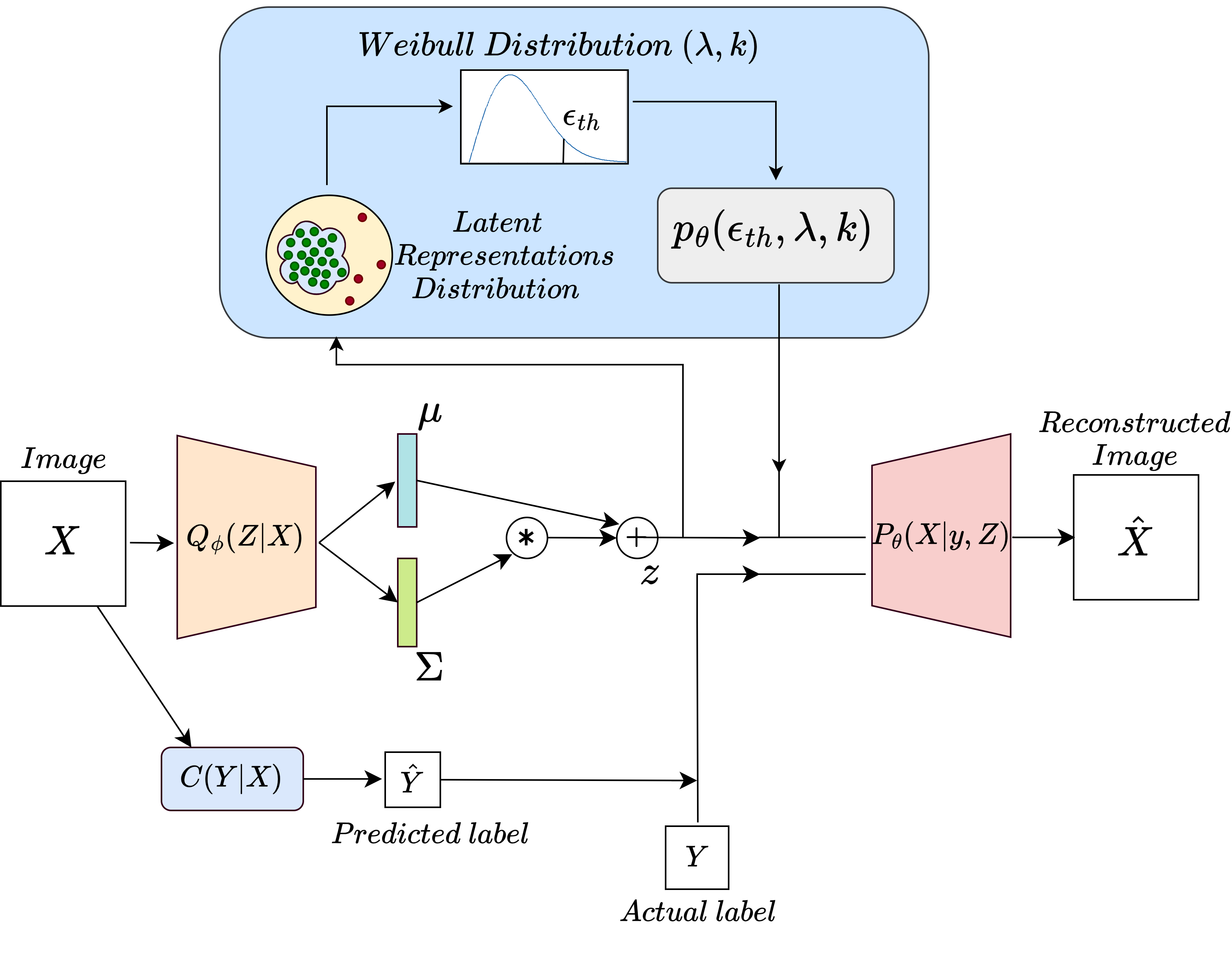}
\caption{Working of CSSAD-VAE.}
\label{fig:vae_intuition}
\centering
\end{figure}

As previously mentioned, Variational Autoencoders (VAEs) are a generative method often used for both semi-supervised anomaly detection and continual anomaly detection \cite{zhang2019semi},\cite{autoencoder_continual_anomaly_detection}. See Figure \ref{fig:vae_intuition} for a diagram illustrating how a VAE works on an intuitive level. 

Let \(x \in X_l, y \in Y_l, x_u \in X_u\).  

\begin{enumerate}
    \item An input \(x_u\) is passed through a neural network with decreasing dimensionality size to produce two vectors, the mean vector, \(\mu\), and the standard deviation, \(\sigma\).
    \item We then use a reparameterization trick to sample from these prior isotropic normal distributions into the latent space, \(z\). 
    \item A classifier is used to classify input \(x_u\)  to a particular class. This is then used to regularise the latent space into a clustered multi-variate normal distribution.
    \item It is then possible to sample from the latent space \(z\) as it is a distribution. After taking a sample \(z'\), this becomes the input to the decoder which is another neural network with the reversed architecture of the encoder. This aims to reproduce the original sample by minimising the difference between the original sample and the output \(x'\). 
\end{enumerate}

In the case where labelled inputs \(x\) and \(y\) are passed to the encoder, the process is the same with the omission of step 3 since classification of the input \(x\) is not required. 

Once the VAE is trained, it is possible to sample from the latent space, z, without needing to process anything through the encoder first, which makes the VAE computationally efficient for novel data generation. 

There are therefore three separate parts to the objective and consequently, loss of the VAE. The first is to optimise the encoder to match the latent space to a prior isotropic normal (KL divergence). The next is to optimise the decoder to minimise the reconstruction error of original input, \(x\) to output \(x'\) (reconstruction loss). Finally, the classifier is to be optimised to to maximise classification accuracy of samples \(x\) when labels are not present (classifier loss).

The reconstruction loss part of the loss function is the Binary Cross Entropy Loss as suggested for use originally by Kingma \textit{et al}. \cite{kingma2019introduction} between an input \(x\), and its reconstruction \(x'\).

In each case, variational inference (VI) \cite{vi_def} is performed to calculate the Evidence Lower Bound (ELBO) \cite{elbo_def} of a single data point (x,y) as this is often intractable for complex distributions. A more detailed derivation of the loss functions used within a VAE can be found here \cite{odaibo2019tutorial}\cite{kingma2019introduction}. 

Given a probabilistic encoder, \(q_\phi(z|x)\) and classifier \(q_\phi(y|x,z)\) with shared parameters \(\phi\), and probabilistic decoder \(p_\theta(x|y, z)\) with parameters \(\theta\), we can define:\\

\begin{align*}
\log p_{\theta}({\bf x}) &\geq E_{q_\phi(y, {\bf z}|{\bf x})}\bigg[
    \log p_{\theta}({\bf x}|y, {\bf z}) + \log p_{\theta}(y)
      + \log p_{\theta}({\bf z}) - \log q_\phi(y, {\bf z}|{\bf x})
\bigg] \\ 
 & = E_{q_\phi(y|{\bf x})}\bigg[
   E_{q_\phi({\bf z}|{\bf x})}\big[
    \log p_{\theta}({\bf x}|y, {\bf z}) + K_1
    + \log p_{\theta}({\bf z})  - \log q_\phi(y|{\bf x}) - \log q_\phi({\bf z}|{\bf x})
   \big]
\bigg] \\
 & = E_{q_\phi(y|{\bf x})}\bigg[
   E_{q_\phi({\bf z}|{\bf x})}\big[
    \log p_{\theta}({\bf x}|y, {\bf z})
   \big]
    + K_1
    - KL[q_{\phi}({\bf z}|{\bf x})||p_{\theta}({\bf z})]
    - \log q_{\phi}(y|{\bf x})
\bigg] \\
&= E_{q_\phi(y|{\bf x})} \big[ -\mathcal{L}({\bf x}, y)
    - \log q_{\phi}(y|{\bf x})
\big] \\ 
&= \sum_y \big[ q_\phi(y|{\bf x})(-\mathcal{L}({\bf x}, y))
    - q_\phi(y|{\bf x}) \log q_\phi(y|{\bf x}) \big] \\
&= \sum_y q_\phi(y|{\bf x})(-\mathcal{L}({\bf x}, y))
    + \mathcal{H}(q_\phi(y|{\bf x})) \\
\end{align*}
where,
\begin{equation}
\mathcal{L}(x,y) =  KL[q_\phi({\bf z}|x)||p_\theta({\bf z})] + K_1 - E_{q_\phi(y|x)}[\log p_\theta(x|y,\bf z)]
\end{equation}
\begin{equation}
\mathcal{H}(q_\phi(y|x)) = - q_\phi(y|x) \log q_\phi(y|x)
\end{equation}
\begin{equation}
\mathcal{L}_{labelled} = \sum_{(x,y) \in {D}}[\mathcal{L}(x,y) - \alpha \log q_\phi(y|x)]
\end{equation}
\begin{equation}
\mathcal{L}_{unlabelled} = \sum_{x } \bigg[ \sum_y q_\phi(y|x)(\mathcal{L}(x,y) - \mathcal{H}(q_\phi(y|x))\bigg]
\end{equation}
\begin{equation}
\mathcal{L}_{total} = \mathcal{L}_{unlabelled} + \mathcal{L}_{labelled}
\end{equation}

\subsubsection{Variations}

One improvement that was made to the base M2 Model  is including a Beta hyperparameter for the KL divergence term in \(L(x,y)\). The updated equation for \(L(x,y)\) can be seen below, where all other parts of the loss function remain unchanged. 

\begin{equation}
\mathcal{L}(x,y) =  \textcolor{red}{\beta} * KL[q_\phi(z|x)||p_\theta(z)] + K_1 - E_{q_\phi(y|x)}[log p_\theta(x|y,z)]
\end{equation}
This was first suggested by Higgins \textit{et al}. \cite{higgins2016beta} and has been shown to balance latent channel capacity and independence constraints with reconstruction accuracy. Another such potential improvement to the baseline model not implemented involves incorporating Ladder variational autoencoders \cite{sonderby2016ladder} into the M2 model.  

\subsection{Continual Learning Approach}

\textbf{Generative Replay With Outlier Rejection}

The main disadvantage of a VAE for data generation is that due to the very nature of the latent space being a distribution, one cannot control the quality of data generation. Building on the low-cost and efficient generative abilities of the VAE, Mundt \textit{et al.} \cite{mundt2022unified} came up with the idea to model how much of a statistical inlier or outlier a generated datapoint is in comparison to the class mean in the latent space. After setting an acceptable outlier threshold, it is then possible to sample from the latent space with outlier rejection. This enables representative sampling of generated datapoints to be used as a generative replay method for continual learning.

Mundt \textit{et al.} \cite{mundt2022unified} propose to regard a sample as a statistical outlier if its distance from the classes latent mean is extreme in comparison to the majority of correctly predicted instances. This is equivalent to a sample falling into a low density zone within the aggregate posterior for the latent space. However, in the case of anomaly detection, where only two classes are present, it is only possible to accurately estimate the latent mean of the normal class. The reason for this is that all of the anomalies may be different - and therefore high-quality replay of each cluster of anomalies becomes impossible for a Weibull distribution to model. 

The latent mean distance for the normal class is defined as: \\
\begin{equation}
    \Delta_0 \equiv {f_d(z_0, E_{q_\theta(\bar{z}|x^{(m)})}[z])_{m \in M_0}}
\end{equation}
where \(M_0\) is the set of correctly identified normal instances in a given experience and \(f_d\) represents a choice of distance metric, chosen to be cosine distance for these experiments. 

The set of distances to the latent mean are estimated by using a per-class Weibull distribution. The sample outlier probability can then be estimated using the CDF of the Weibull model shown below. 

\begin{equation}
w_p(z) \equiv {min(1- exp(-\frac{|f_d(\bar{z},z) - \tau|}{\lambda})^\kappa)}
\end{equation}

where \begin{math}
p_0(z) = (\tau, \kappa, \lambda)
\end{math} is a univariate heavy-tailed Weibull model trained on the normal class.

If this is below a rejection probability which is determined through using a validation set, then the sample is rejected. Since all of this is happening before the sample z is processed through the decoder, it is computationally efficient. Through representative generative replay, only high-quality samples are replayed which help to mitigate catastrophic forgetting. 

\subsubsection{Other Methods Employed}

\textbf{Naive} \\
The lower bound for continual learning methods will be naive training in which no continual learning method is implemented and the model trains sequentially on the disjoint datasets without a strategy to mitigate catastrophic forgetting. 

\textbf{Joint Training} \\
The upper bound for each experiment will be joint training. This is where the model is trained on the entire dataset at once as is normal in supervised or unsupervised training. This constitutes the upper bound for what is possible for a model as no continual learning method is necessary to mitigate catastrophic forgetting. In this CSSL setting, it is the equivalent of purely SSL. 

\textbf{EWC} \\
Elastic Weight Consolidation was introduced by Kirkpatrick \textit{et al.} in 2017 \cite{ewc_cl}. This selectively constrains the model weights which are most important to a specific task. In the domain-incremental setting, a trade-off between plasticity and stability is desirable such that the model adapts to the changing data distribution, but does not do so too quickly so as to forget the previously seen data distribution. This is going to be implemented as a baseline method in which to compare other continual learning methods against. 

\subsection{Approach to Anomaly Detection}

An and Cho \cite{an2015variational} proposed using the reconstruction probability of VAEs as a means of anomaly detection. In their paper, they demonstrate how to calculate the reconstruction probability of a sample, and if it is below a certain threshold, it is deemed to be anomalous. However, the ELBO which is already calculated for the VAE loss, can also be used to approximate the reconstruction probability as in \cite{autoencoder_continual_anomaly_detection}, which can prove to give better results under certain circumstances. Due to the ease of implementation, this will be used as a baseline anomaly detection method, acknowledging that more research should be done in this area to potentially improve results within the benchmarks being set out. 

Another natural choice for anomaly detection would be the outlier rejection probability \cite{mundt2022unified}. This could be used in a similar way to the reconstruction probability or ELBO which will use the AUC calculated over the test set to find the optimal threshold for these particular metrics which maximises the AUC. This paper does not employ the outlier rejection probability due to the implementation of the probabilistic encoder in line with Kingma \textit{et al}.  \cite{kingma2019introduction}. In this paper, the encoder is conditioned upon \(y\) such that to encode an input, \(x\), its label \(y\) is required. This is in direct opposition to the assumptions of Mundt \textit{et al}. \cite{mundt2022unified} that an input \(x\) can be encoded into the latent space without its label \(y\). In breaking this assumption, it is not possible to calculate the distance from each latent mean as it is already conditioned on \(y\). However, this will be left to future researchers to explore the effects of different approaches to anomaly detection within the scope of SSAD.

\section{Results}

\subsection{Benchmark Datasets}

In order to empirically validate CSAD methods that are being employed, artificial anomaly detection datasets are used in order to allow ablation studies with varying levels of labelled data as well as labelled and unlabelled anomalies within the training dataset.

The \textbf{MNIST} dataset, accessed \href{http://yann.lecun.com/exdb/mnist/}{here}, is a collection of 70,000 handwritten digits split into 60,000 training and 10,000 test observations. It has 10 classes which represent the numbers from 0 to 9 \cite{deng2012mnist}.

The \textbf{CIFAR-10} dataset, accessed \href{https://www.cs.toronto.edu/~kriz/cifar.html}{here}, is a subset of the tiny images dataset \cite{tiny_images} containing 60,000 images of 10 classes split into 50,000 training and 10,000 test observations \cite{krizhevsky2009learning}. 

The \textbf{Fashion MNIST} dataset, accessed \href{https://github.com/zalandoresearch/fashion-mnist} {here}, consists of 70,000 clothing articles split into 60,000 training and 10,000 test observations associated with a label from 10 classes \cite{xiao2017fashion}. 

All datasets are flattened, scaled to between 0 and 1, and one hot encoding is applied to the targets. For MNIST and Fashion MNIST, data augmentation using AugMix \cite{hendrycks2019augmix} is applied to the normal labelled data to double the amount of normal labelled training instances available. Since these datasets are all intended for supervised classification in a non-continual setting, there have to be choices made about how the data should be split across different experiences. This depends on five different attributes of the dataset: 1. the overall percentage of labelled normal data in the dataset (\(\alpha\)); 2. the spread of labelled normal data across experiences (\(\beta\)); 3. the percentage of labelled anomalies within an experience  (\(\gamma\)); 4. the percentage of unlabelled anomalies within an experience (\(\zeta\)); 5. the anomalous classes within an experience (\(\lambda\)).

\begin{table}[h]
    \centering
    \caption{Definitions of Variables}
    \begin{tabular}{@{}ll@{}}
        \toprule
        \textbf{Variable} & \textbf{Definition} \\ \midrule
        $\alpha$ & Overall percentage of labelled normal data in the dataset \\
        $\beta$ & Spread of labelled normal data across experiences \\
        $\gamma$ & Percentage of labelled anomalies within an experience \\
        $\zeta$ & Percentage of unlabelled anomalies within an experience \\
        $\lambda$ & Anomalous classes within an experience \\ 
        \bottomrule
    \end{tabular}
    \label{tab:variables}
\end{table}

Following from this, one class is chosen as the normal class as by Ruff \textit{et al}. \cite{ruff2019deep} in their extensive experiments on synthetic anomaly detection datasets. All of the other classes are set as anomalous. Once the normal class is chosen, labelled normal data is sampled randomly until the threshold of \(\alpha\) is met. Then, for each experience, this labelled normal data is spread according to the parameter \(\beta\). Finally, labelled and unlabelled data are added to each experience in the percentage amounts as specified by \(\gamma\) and \(\zeta\). The classes within an experience are randomly sampled from available classes for that particular experience (e.g. [0,1,2] for experience 1 will sample randomly from the pools of these classes before the anomaly class transformation is applied). The classes are set up in such a way to introduce new anomalous classes over time indicating the domain-incremental data distribution shift which will instigate catastrophic forgetting. The classes present in each experience are an additional parameter which could be further tested through ablation studies, but were outside the scope of this paper. Whilst there are undoubtedly drawbacks of using artificial anomaly detection datasets such as MNIST \cite{deng2012mnist}, CIFAR-10 \cite{krizhevsky2009learning}, and Fashion MNIST \cite{xiao2017fashion}, the obvious advantage is the ability to create many different artificial subsets on which to test the performance of anomaly detection models in extensive ablation studies.

\subsection{Experiments}
There are 3 different data streams within an experiment, the training, the testing stream, and the validation stream. In order to create the experiences which make up the data streams, the initial training and testing data is used. However, due to the sampling procedure, the training streams may differ in size between experiments due to the addition or withholding of labelled and unlabelled anomalous data. 

\textbf{Training stream} \\
Firstly, the anomalous classes are separated from the training data and the remaining data is split into labelled and unlabelled data using a stratified split based on \(\alpha\). Then, for an individual experience, labelled normal data is sampled randomly based on \(\beta\). Anomalies are then added into the labelled and unlabelled data to meet the percentage of anomalous data within the respective dataset based on \(\gamma\) and \(\zeta\) along with the available underlying anomalous classes \(\lambda\). The result is that there are a number of labelled and unlabelled experiences which make up the training data stream. 

\textbf{Validation stream} \\
A stratified split is taken from the labelled data stream to make up the validation stream. This is normally created as 10\% of the training stream using a stratified split from the training experiences to keep the distribution of labelled normal data and anomalies.

\textbf{Testing stream} \\
The testing stream is created by carrying out a stratified split across the testing data to create equally sized experiences which represent the original distribution of data. This split is based on the original distribution of the targets of the data before anomalous and normal class transformations are applied. 

\subsection{Evaluation Metrics}

The chosen evaluation metric for experiments is AUC. This is defined by Bradley \cite{bradley1997use} as the area under the receiver operator curve (ROC) which plots the False positive rate against the True positive rate. The AUC measures how well a model can separate between two classes and as such is the obvious choice within anomaly detection. This is further reinforced through it consistently being the metric of choice within anomaly detection research \cite{samuel2021svd}\cite{ruff2019deep}\cite{autoencoder_continual_anomaly_detection}. 

\subsection{Main Results}
\begin{table}[]
\centering
\caption{\small AUC-ROC Results by Dataset: The AUCROC scores achieved by various continual learning strategies — EWC (Elastic Weight Consolidation), Joint, Naive, and OR (Outlier Rejection)—across three benchmark datasets: MNIST, Fashion MNIST, and CIFAR-10. \textbf{Bold} for the best result and \underline{underline} for the second-best result}
 \label{tab: main results}
\resizebox{0.8\textwidth}{!}{  
\begin{tabular}{c|ccc}

\toprule
\multirow{2}{*}{\textbf{Method}} & \multicolumn{3}{c}{\textbf{Dataset}} \\ 
\cmidrule{2-4} 
 & \textbf{MNIST} & \textbf{Fashion MNIST} & \textbf{CIFAR-10} \\
\midrule
EWC   & 0.646 & 0.521 & 0.544 \\
Joint & \underline{0.655} & 0.401 & \textbf{0.549} \\
Naive & 0.645 & \underline{0.555} & \textbf{0.549} \\

\hline

\textbf{Outlier Rejection (OR) (ours)}    & \textbf{0.690} & \textbf{0.581} & \underline{0.546} \\
\bottomrule

\end{tabular}
}
\end{table}

The results in the table demonstrate that the Outlier Rejection (OR) method outperforms other continual learning strategies across most datasets, achieving the highest AUC-ROC scores on MNIST (0.690) and Fashion MNIST (0.581). This highlights the effectiveness of OR in these tasks. While OR does not achieve the best score on CIFAR-10, it remains competitive, with a score of 0.546, slightly trailing Joint and Naive methods, which both achieve 0.549. These findings suggest that OR is particularly well-suited for simpler datasets like MNIST and Fashion MNIST, while its performance on more complex datasets like CIFAR-10 is comparable but not superior to other methods.

\setlength{\tabcolsep}{6pt}

\begin{table}[h!]
\centering
\caption{AUCROC Scores for Continual Learning Strategies on MNIST Across Episodes}
\resizebox{\textwidth}{!}{ 
\begin{tabular}{c|ccccccccccccccc}
\toprule
\multirow{2}{*}{\textbf{Method}} & \multicolumn{15}{c}{\textbf{Episode Number}} \\
\cmidrule(l){2-16}
& \textbf{1} & \textbf{2} & \textbf{3} & \textbf{4} & \textbf{5} & \textbf{6} & \textbf{7} & \textbf{8} & \textbf{9} & \textbf{10} & \textbf{11} & \textbf{12} & \textbf{13} & \textbf{14} & \textbf{15} \\
\midrule
EWC   & 0.69 & 0.64 & 0.64 & 0.67 & 0.65 & 0.70 & 0.65 & 0.75 & 0.51 & 0.71 & 0.58 & 0.68 & 0.68 & 0.63 & 0.51 \\
Joint & \textbf{0.73} & 0.63 & 0.67 & 0.71 & 0.65 & 0.69 & 0.69 & 0.82 & 0.53 & 0.65 & 0.59 & 0.70 & 0.67 & 0.58 & 0.52 \\
Naive & 0.69 & 0.63 & 0.68 & 0.60 & 0.66 & 0.66 & 0.66 & 0.72 & 0.64 & 0.65 & 0.61 & 0.69 & 0.64 & 0.61 & 0.54 \\
OR    & 0.60 & \textbf{0.64} & 0.67 & \textbf{0.78} & \textbf{0.76} & 0.67 & \textbf{0.77} & \textbf{0.84} & 0.59 & 0.71 & 0.66 & \textbf{0.73} & 0.68 & 0.64 & \textbf{0.61} \\
\bottomrule
\end{tabular}
}
\end{table}

\begin{table}[h!]
\centering
\caption{AUCROC Scores for Continual Learning Strategies on Fashion MNIST Across Episodes}
\resizebox{\textwidth}{!}{ 
\begin{tabular}{c|ccccccccccccccc}
\toprule

\multirow{2}{*}{\textbf{Method}} & \multicolumn{15}{c}{\textbf{Episode Number}} \\
\cmidrule(l){2-16}
& \textbf{1} & \textbf{2} & \textbf{3} & \textbf{4} & \textbf{5} & \textbf{6} & \textbf{7} & \textbf{8} & \textbf{9} & \textbf{10} & \textbf{11} & \textbf{12} & \textbf{13} & \textbf{14} & \textbf{15} \\
\midrule

EWC   & 0.40 & \textbf{0.55} & 0.58 & 0.73 & 0.54 & 0.39 & 0.48 & 0.65 & 0.55 & 0.56 & 0.44 & 0.52 & 0.57 & 0.56 & 0.29 \\
Joint & 0.49 & 0.35 & 0.44 & 0.43 & 0.38 & 0.39 & 0.43 & 0.64 & 0.33 & 0.39 & 0.33 & 0.40 & 0.35 & 0.38 & 0.29 \\
Naive & 0.69 & 0.54 & 0.55 & 0.58 & 0.53 & 0.55 & 0.61 & 0.67 & 0.53 & 0.55 & 0.56 & 0.55 & 0.56 & 0.57 & 0.29 \\
OR    & \textbf{0.70} & 0.53 & \textbf{0.59} & \textbf{0.75} & \textbf{0.78} & \textbf{0.62} & \textbf{0.73} & 0.56 & 0.45 & \textbf{0.60} & 0.46 & \textbf{0.60} & 0.55 & 0.49 & \textbf{0.30} \\
\bottomrule
\end{tabular}
}
\end{table}

\begin{table}[h!]
\centering
\caption{AUCROC Scores for Continual Learning Strategies on CIFAR-10 Across Episodes}
\resizebox{\textwidth}{!}{ 
\begin{tabular}{c|ccccccccccccccc}
\toprule
\multirow{2}{*}{\textbf{Method}} & \multicolumn{15}{c}{\textbf{Episode Number}} \\
\cmidrule(l){2-16}
& \textbf{1} & \textbf{2} & \textbf{3} & \textbf{4} & \textbf{5} & \textbf{6} & \textbf{7} & \textbf{8} & \textbf{9} & \textbf{10} & \textbf{11} & \textbf{12} & \textbf{13} & \textbf{14} & \textbf{15} \\
\midrule

EWC   & 0.54 & 0.54 & 0.58 & 0.59 & 0.54 & 0.54 & 0.53 & 0.52 & 0.58 & 0.51 & 0.56 & 0.55 & 0.59 & 0.55 & 0.44 \\
Joint & 0.59 & 0.56 & 0.58 & 0.51 & 0.52 & 0.55 & 0.59 & \textbf{0.62} & 0.54 & 0.56 & 0.55 & 0.52 & 0.55 & 0.56 & 0.44 \\
Naive & 0.55 & \textbf{0.57} & \textbf{0.60} & 0.58 & 0.54 & \textbf{0.59} & 0.52 & 0.53 & 0.52 & 0.55 & 0.56 & 0.53 & \textbf{0.60} & 0.54 & \textbf{0.46} \\
OR    & \textbf{0.61} & 0.53 & 0.57 & 0.53 & \textbf{0.56} & 0.52 & \textbf{0.55} & 0.48 & 0.53 & 0.55 & \textbf{0.64} & 0.53 & 0.55 & \textbf{0.60} & 0.44 \\
\bottomrule
\end{tabular}
}
\end{table}

\subsection{Ablation Studies}

There are four different ablation studies that will be carried out across the different datasets, consisting of a total of 15 different experiments being run. Throughout these ablation studies, please refer to the definitions of \(\alpha\), \(\beta\), \(\gamma\), and \(\zeta\) as laid out in \ref{tab:variables}.

\textbf{1. Varying labelled data percentage \(\alpha\)}

Within this ablation study, \(\alpha\) is varied within the datasets from 5 to 20\% as shown in experiments 1  to 3 which can be found in Appendix \ref{experiment_1_appendix}, \ref{experiment_2_appendix}, and \ref{experiment_3_appendix}. Please note that for all Appendix plots, an additional black and white printable version is made available. 

Default values for the other four data-varying parameters can be seen below: 
\begin{itemize}
    \item \(\beta\) - spread of labelled data is equal across experiences
    \item \(\gamma\) - percentage of labelled anomalies in an experience is 5 \%
    \item \(\zeta\) - percentage of unlabelled anomalies in each experience 0\%
    \item \(\lambda\) - all classes are present in each experience
\end{itemize}

\textbf{2. Varying percentage of labelled data in each experience, \(\beta\)}

This ablation study varies \(\beta\) across different experiences, looking at varying levels of skew towards the first and last experiences. An example experiment can be seen below, and the remaining experiments can be found in experiments 4  to 7 which can be found in Appendix \ref{experiment_4_appendix}, \ref{experiment_5_appendix},\ref{experiment_6_appendix} and \ref{experiment_7_appendix}.  

\textbf{3. Varying percentage of labelled anomalies in each experience, \(\gamma\)} \\
In Ablation study 3, the percentage of labelled anomalies within each experience \(\gamma\) is explored. This is kept fixed across all experiences, and varying the percentage of labelled anomalous data between experiences will be left for future study. This study represents experiments 8 through 11, where individual results for each experiment can be viewed in \ref{experiment_8_appendix}, \ref{experiment_9_appendix}, \ref{experiment_10_appendix}, and \ref{experiment_11_appendix}. 

The default value for \(\beta\) is set to [0.2, 0.2, 0.2, 0.2, 0.2], and the other hyperparamter values remain the same. 

\textbf{4. Varying percentage of unlabelled anomalies in each experience, \(\zeta\)}

In this study, the percentage of unlabelled anomalies will be varied across each experience. This will likely negatively impact the ability of the VAE to reconstruct the normal class as there may be similarity between the normal class and the unlabelled anomalous sample which may therefore form a similar cluster within the latent space. This ablation study can be found within experiments 12 to 15 where experimental setup, along with results can be found in Appendix \ref{experiment_12_appendix}, \ref{experiment_13_appendix},\ref{experiment_14_appendix} and \ref{experiment_15_appendix}. 

\section{Ablation Results}

The results for the ablation studies, each run for 10 training epochs with early stopping, are presented below. Please refer to \ref{tab:variables} for definitions of \(\alpha\), \(\beta\), \(\gamma\), and \(\zeta\).

\textbf{Ablation 1}
As can be seen in Figure \ref{experiment_1_appendix}, for MNIST, the AUC for outlier rejection often outperforms the baseline, Naive, and in most cases outperforms EWC. This indicates that if a VAE is able to learn the normal class sufficiently well during training, it is entirely possible to replay high quality data that is representative of the normal class. All of this is in spite of the fact that the classifier being used for the normal class are simple convolutional layers, which could be greatly improved upon with better architectures. 


\begin{figure}[h]
    \centering
    \begin{subfigure}{0.6\textwidth}
        \centering
        \begin{tikzpicture}
            \begin{axis}[
                height=6cm,
                width=10cm,
                ybar=0.1pt,
                bar width=0.4cm,
                enlarge x limits=0.25,
                symbolic x coords={MNIST, CIFAR10, Fashion MNIST},
                xtick=data,
                ylabel={AUC Score},
                ymajorgrids=true,
                grid style=dashed,
                ymin=0.3, 
                ymax=0.8, 
                ytick={0.4, 0.5, 0.6, 0.7, 0.8},
                legend image code/.code={
                \draw [#1] (0cm,0.1cm) rectangle (0.15cm,0.15cm); },
                x tick label style={rotate=0, anchor=north},
                legend style={at={(0.5,1.05)}, anchor=south, legend columns=2, nodes={scale=0.8, transform shape}},
                label style={font=\footnotesize},
                tick label style={font=\footnotesize}  
            ]
            \addplot[ybar, fill=blue] coordinates {(MNIST,0.6906) (CIFAR10,0.5538) (Fashion MNIST,0.6864)};
            \addplot[ybar, fill=orange] coordinates {(MNIST,0.7281) (CIFAR10,0.5896) (Fashion MNIST,0.4916)};
            \addplot[ybar, fill=green] coordinates {(MNIST,0.6942) (CIFAR10,0.5444) (Fashion MNIST,0.4027)};
            \addplot[ybar, fill=purple] coordinates {(MNIST,0.6043) (CIFAR10,0.6102) (Fashion MNIST,0.6973)};

            \addlegendimage{ybar,fill=blue}
            \addlegendentry{Naive}
            \addlegendimage{ybar,fill=orange}
            \addlegendentry{Joint Training}
            \addlegendimage{ybar,fill=green}
            \addlegendentry{EWC}
            \addlegendimage{ybar,fill=purple}
            \addlegendentry{Outlier Rejection}
            \end{axis}
        \end{tikzpicture}
        \caption{\textbf{AUC Scores (Exp 1)}}
    \end{subfigure}
    \caption{AUC Scores for Different Methods on MNIST, CIFAR10, and Fashion MNIST}
    \label{experiment_1_appendix}
\end{figure}
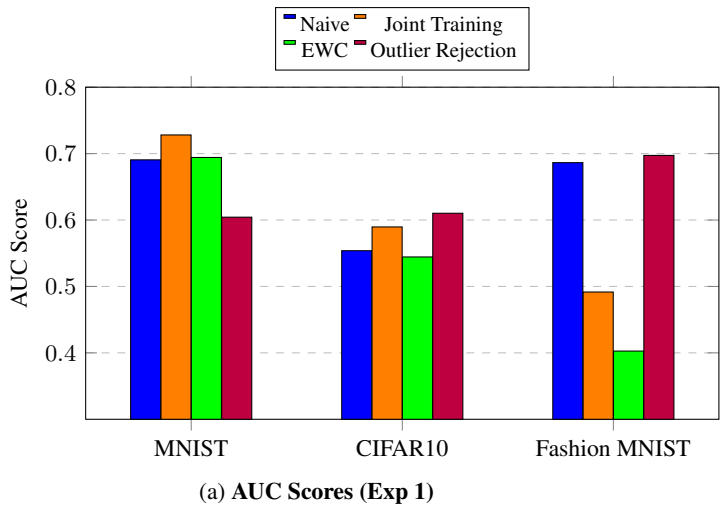

\begin{figure}[h]
    \centering
    \begin{subfigure}{0.6\textwidth}
        \centering
        \begin{tikzpicture}
            \begin{axis}[
                height=6cm,
                width=10cm,
                ybar=0.1pt,
                bar width=0.4cm,
                enlarge x limits=0.25,
                symbolic x coords={MNIST, CIFAR10, Fashion MNIST},
                xtick=data,
                ylabel={AUC Score},
                ymajorgrids=true,
                grid style=dashed,
                ymin=0.3, 
                ymax=0.8, 
                ytick={0.4, 0.5, 0.6, 0.7, 0.8},
                legend image code/.code={
                \draw [#1] (0cm,0.1cm) rectangle (0.15cm,0.15cm); },
                x tick label style={rotate=0, anchor=north},
                legend style={at={(0.5,1.05)}, anchor=south, legend columns=2, nodes={scale=0.8, transform shape}},
                label style={font=\footnotesize},
                tick label style={font=\footnotesize}  
            ]
            \addplot[ybar, fill=blue] coordinates {(MNIST,0.6264) (CIFAR10,0.5686) (Fashion MNIST,0.5434)};
            \addplot[ybar, fill=orange] coordinates {(MNIST,0.6317) (CIFAR10,0.5624) (Fashion MNIST,0.3498)};
            \addplot[ybar, fill=green] coordinates {(MNIST,0.6358) (CIFAR10,0.5388) (Fashion MNIST,0.5509)};
            \addplot[ybar, fill=purple] coordinates {(MNIST,0.6438) (CIFAR10,0.5255) (Fashion MNIST,0.5303)};

            \addlegendimage{ybar,fill=blue}
            \addlegendentry{Naive}
            \addlegendimage{ybar,fill=orange}
            \addlegendentry{Joint Training}
            \addlegendimage{ybar,fill=green}
            \addlegendentry{EWC}
            \addlegendimage{ybar,fill=purple}
            \addlegendentry{Outlier Rejection}
            \end{axis}
        \end{tikzpicture}
        \caption{\textbf{AUC Scores (Exp 2)}}
    \end{subfigure}
    \caption{AUC Scores for Different Methods on MNIST, CIFAR10, and Fashion MNIST}
    \label{experiment_2_appendix}
\end{figure}

\begin{figure}[h]
    \centering
    \begin{subfigure}{0.6\textwidth}
        \centering
        \begin{tikzpicture}
            \begin{axis}[
                height=6cm,
                width=10cm,
                ybar=0.1pt,
                bar width=0.4cm,
                enlarge x limits=0.25,
                symbolic x coords={MNIST, CIFAR10, Fashion MNIST},
                xtick=data,
                ylabel={AUC Score},
                ymajorgrids=true,
                grid style=dashed,
                ymin=0.3, 
                ymax=0.8, 
                ytick={0.4, 0.5, 0.6, 0.7, 0.8},
                legend image code/.code={
                \draw [#1] (0cm,0.1cm) rectangle (0.15cm,0.15cm); },
                x tick label style={rotate=0, anchor=north},
                legend style={at={(0.5,1.05)}, anchor=south, legend columns=2, nodes={scale=0.8, transform shape}},
                label style={font=\footnotesize},
                tick label style={font=\footnotesize}  
            ]
\addplot[ybar, fill=blue] coordinates {(MNIST,0.6824) (CIFAR10,0.6047) (Fashion MNIST,0.5451)};
\addplot[ybar, fill=orange] coordinates {(MNIST,0.6672) (CIFAR10,0.5769) (Fashion MNIST,0.4352)};
\addplot[ybar, fill=green] coordinates {(MNIST,0.6378) (CIFAR10,0.5824) (Fashion MNIST,0.5806)};
\addplot[ybar, fill=purple] coordinates {(MNIST,0.6731) (CIFAR10,0.5748) (Fashion MNIST,0.5940)};

            \addlegendimage{ybar,fill=blue}
            \addlegendentry{Naive}
            \addlegendimage{ybar,fill=orange}
            \addlegendentry{Joint Training}
            \addlegendimage{ybar,fill=green}
            \addlegendentry{EWC}
            \addlegendimage{ybar,fill=purple}
            \addlegendentry{Outlier Rejection}
            \end{axis}
        \end{tikzpicture}
        \caption{\textbf{AUC Scores (Exp 3)}}
    \end{subfigure}
    \caption{AUC Scores for Different Methods on MNIST, CIFAR10, and Fashion MNIST}
    \label{experiment_3_appendix}
\end{figure}

When varying the percentage of labelled data overall in the dataset, there is an obvious upward trend in performance as the \(\alpha\) is increased. This indicates that the M2 model is correctly utilising the labelled data in the semi-supervised portion of this training setting. However, There were issues in joint training which meant that it was not in fact the upper bound for all experiments. This is especially apparent in the Fashion MNIST dataset.

However, for a dataset such as CIFAR-10, the AUC is consistently low, even for joint training, indicating that convolutional layers are required in the encoder and decoder as simple dense layers are unlikely to be able to correctly reconstruct CIFAR-10 images. This also explains why outlier rejection consistently performs worse than EWC in this instance as the model can not accurately reproduce high-quality images for generative replay to mitigate catastrophic forgetting. In their paper, Urban \textit{et al}. \cite{urban2016deep} conclude that MLPs cannot rival the accuracy of CNNs when training on CIFAR-10 and CIFAR-100 \cite{krizhevsky2009learning}. Whilst a convolutional neural network (CNN) was implemented for the classifier part of the network, it was not implemented in the encoder and decoder due to time constraints in running all of the experiments.      

\textbf{Ablation 2}

\begin{figure}[h]
    \centering
    \begin{subfigure}{0.6\textwidth}
        \centering
        \begin{tikzpicture}
            \begin{axis}[
                height=6cm,
                width=10cm,
                ybar=0.1pt,
                bar width=0.4cm,
                enlarge x limits=0.25,
                symbolic x coords={MNIST, CIFAR10, Fashion MNIST},
                xtick=data,
                ylabel={AUC Score},
                ymajorgrids=true,
                grid style=dashed,
                ymin=0.3, 
                ymax=0.8, 
                ytick={0.4, 0.5, 0.6, 0.7, 0.8},
                legend image code/.code={
                \draw [#1] (0cm,0.1cm) rectangle (0.15cm,0.15cm); },
                x tick label style={rotate=0, anchor=north},
                legend style={at={(0.5,1.05)}, anchor=south, legend columns=2, nodes={scale=0.8, transform shape}},
                label style={font=\footnotesize},
                tick label style={font=\footnotesize}  
            ]
            \addplot[ybar, fill=blue] coordinates {(MNIST,0.6029) (CIFAR10,0.5780) (Fashion MNIST,0.5839)};
            \addplot[ybar, fill=orange] coordinates {(MNIST,0.7096) (CIFAR10,0.5142) (Fashion MNIST,0.4324)};
            \addplot[ybar, fill=green] coordinates {(MNIST,0.6694) (CIFAR10,0.5883) (Fashion MNIST,0.7306)};
            \addplot[ybar, fill=purple] coordinates {(MNIST,0.7772) (CIFAR10,0.5334) (Fashion MNIST,0.7497)};
            
            \addlegendimage{ybar,fill=blue}
            \addlegendentry{Naive}
            \addlegendimage{ybar,fill=orange}
            \addlegendentry{Joint Training}
            \addlegendimage{ybar,fill=green}
            \addlegendentry{EWC}
            \addlegendimage{ybar,fill=purple}
            \addlegendentry{Outlier Rejection}
            \end{axis}
        \end{tikzpicture}
        \caption{\textbf{AUC Scores (Exp 4)}}
    \end{subfigure}
    \caption{AUC Scores for Different Methods on MNIST, CIFAR10, and Fashion MNIST}
    \label{experiment_4_appendix}
\end{figure}

\begin{figure}[h]
    \centering
    \begin{subfigure}{0.6\textwidth}
        \centering
        \begin{tikzpicture}
            \begin{axis}[
                height=6cm,
                width=10cm,
                ybar=0.1pt,
                bar width=0.4cm,
                enlarge x limits=0.25,
                symbolic x coords={MNIST, CIFAR10, Fashion MNIST},
                xtick=data,
                ylabel={AUC Score},
                ymajorgrids=true,
                grid style=dashed,
                ymin=0.3, 
                ymax=0.8, 
                ytick={0.4, 0.5, 0.6, 0.7, 0.8},
                legend image code/.code={
                \draw [#1] (0cm,0.1cm) rectangle (0.15cm,0.15cm); },
                x tick label style={rotate=0, anchor=north},
                legend style={at={(0.5,1.05)}, anchor=south, legend columns=2, nodes={scale=0.8, transform shape}},
                label style={font=\footnotesize},
                tick label style={font=\footnotesize}  
            ]
\addplot[ybar, fill=blue] coordinates {(MNIST,0.6599) (CIFAR10,0.5427) (Fashion MNIST,0.5335)};
\addplot[ybar, fill=orange] coordinates {(MNIST,0.6456) (CIFAR10,0.5208) (Fashion MNIST,0.3833)};
\addplot[ybar, fill=green] coordinates {(MNIST,0.6478) (CIFAR10,0.5363) (Fashion MNIST,0.5406)};
\addplot[ybar, fill=purple] coordinates {(MNIST,0.7585) (CIFAR10,0.5566) (Fashion MNIST,0.7795)};

            \addlegendimage{ybar,fill=blue}
            \addlegendentry{Naive}
            \addlegendimage{ybar,fill=orange}
            \addlegendentry{Joint Training}
            \addlegendimage{ybar,fill=green}
            \addlegendentry{EWC}
            \addlegendimage{ybar,fill=purple}
            \addlegendentry{Outlier Rejection}
            \end{axis}
        \end{tikzpicture}
        \caption{\textbf{AUC Scores (Exp 5)}}
    \end{subfigure}
    \caption{AUC Scores for Different Methods on MNIST, CIFAR10, and Fashion MNIST}
    \label{experiment_5_appendix}
\end{figure}

\begin{figure}[h]
    \centering
    \begin{subfigure}{0.6\textwidth}
        \centering
        \begin{tikzpicture}
            \begin{axis}[
                height=6cm,
                width=10cm,
                ybar=0.1pt,
                bar width=0.4cm,
                enlarge x limits=0.25,
                symbolic x coords={MNIST, CIFAR10, Fashion MNIST},
                xtick=data,
                ylabel={AUC Score},
                ymajorgrids=true,
                grid style=dashed,
                ymin=0.3, 
                ymax=0.8, 
                ytick={0.4, 0.5, 0.6, 0.7, 0.8},
                legend image code/.code={
                \draw [#1] (0cm,0.1cm) rectangle (0.15cm,0.15cm); },
                x tick label style={rotate=0, anchor=north},
                legend style={at={(0.5,1.05)}, anchor=south, legend columns=2, nodes={scale=0.8, transform shape}},
                label style={font=\footnotesize},
                tick label style={font=\footnotesize}  
            ]
\addplot[ybar, fill=blue] coordinates {(MNIST,0.6630) (CIFAR10,0.5890) (Fashion MNIST,0.5465)};
\addplot[ybar, fill=orange] coordinates {(MNIST,0.6871) (CIFAR10,0.5472) (Fashion MNIST,0.3900)};
\addplot[ybar, fill=green] coordinates {(MNIST,0.6974) (CIFAR10,0.5396) (Fashion MNIST,0.3930)};
\addplot[ybar, fill=purple] coordinates {(MNIST,0.6665) (CIFAR10,0.5168) (Fashion MNIST,0.6176)};

            \addlegendimage{ybar,fill=blue}
            \addlegendentry{Naive}
            \addlegendimage{ybar,fill=orange}
            \addlegendentry{Joint Training}
            \addlegendimage{ybar,fill=green}
            \addlegendentry{EWC}
            \addlegendimage{ybar,fill=purple}
            \addlegendentry{Outlier Rejection}
            \end{axis}
        \end{tikzpicture}
        \caption{\textbf{AUC Scores (Exp 6)}}
    \end{subfigure}
    \caption{AUC Scores for Different Methods on MNIST, CIFAR10, and Fashion MNIST}
    \label{experiment_6_appendix}
\end{figure}

\begin{figure}[h]
    \centering
    \begin{subfigure}{0.6\textwidth}
        \centering
        \begin{tikzpicture}
            \begin{axis}[
                height=6cm,
                width=10cm,
                ybar=0.1pt,
                bar width=0.4cm,
                enlarge x limits=0.25,
                symbolic x coords={MNIST, CIFAR10, Fashion MNIST},
                xtick=data,
                ylabel={AUC Score},
                ymajorgrids=true,
                grid style=dashed,
                ymin=0.3, 
                ymax=0.8, 
                ytick={0.4, 0.5, 0.6, 0.7, 0.8},
                legend image code/.code={
                \draw [#1] (0cm,0.1cm) rectangle (0.15cm,0.15cm); },
                x tick label style={rotate=0, anchor=north},
                legend style={at={(0.5,1.05)}, anchor=south, legend columns=2, nodes={scale=0.8, transform shape}},
                label style={font=\footnotesize},
                tick label style={font=\footnotesize}  
            ]
\addplot[ybar, fill=blue] coordinates {(MNIST,0.6590) (CIFAR10,0.5168) (Fashion MNIST,0.6134)};
\addplot[ybar, fill=orange] coordinates {(MNIST,0.6858) (CIFAR10,0.5851) (Fashion MNIST,0.4274)};
\addplot[ybar, fill=green] coordinates {(MNIST,0.6456) (CIFAR10,0.5270) (Fashion MNIST,0.4781)};
\addplot[ybar, fill=purple] coordinates {(MNIST,0.7686) (CIFAR10,0.5512) (Fashion MNIST,0.7293)};

            \addlegendimage{ybar,fill=blue}
            \addlegendentry{Naive}
            \addlegendimage{ybar,fill=orange}
            \addlegendentry{Joint Training}
            \addlegendimage{ybar,fill=green}
            \addlegendentry{EWC}
            \addlegendimage{ybar,fill=purple}
            \addlegendentry{Outlier Rejection}
            \end{axis}
        \end{tikzpicture}
        \caption{\textbf{AUC Scores (Exp 7)}}
    \end{subfigure}
    \caption{AUC Scores for Different Methods on MNIST, CIFAR10, and Fashion MNIST}
    \label{experiment_7_appendix}
\end{figure}

In Figure \ref{experiment_4_appendix}, we can see the results of ablation study 2 which covers experiments 4 through 7. Within this, the distribution of the labelled data across experiences is varied. \(\beta_1\) and \(\beta_2\) represent the experiments in which there is 80\% of the data at either the first or last experience and 5\% otherwise. As expected, we can see that the results perform better with the labelled data in the later experience, especially with the Naive method. The reason for this is that less action is required to mitigate catastrophic forgetting, as seen by the vast improvement of the Naive method in particular, which does nothing to mitigate catastrophic forgetting. However, it is interesting to see that the performance of EWC consistently drops across all datasets - suggesting that the quality of the data in the last experience is not as high as the first. Further experiments would need to be undertaken to confirm this. 

In the experiments representing \(\beta_3\) and \(\beta_4\), the labelled data is exponentially distributed across the experiences. The results again suggest that labelled data being skewed towards the later experiences reduces the impact of continual learning methods. This is seen from the smaller spread of results in \(\beta_3\) in comparison to \(\beta_4\). However, the quality of the data is again put into question where the Naive method increases in AUC from \(\beta_3\) to \(\beta_4\) for Fashion MNIST indicating the data quality in the last experience is higher in \(\beta_4\) than \(\beta_3\). 

\textbf{Ablation 3}

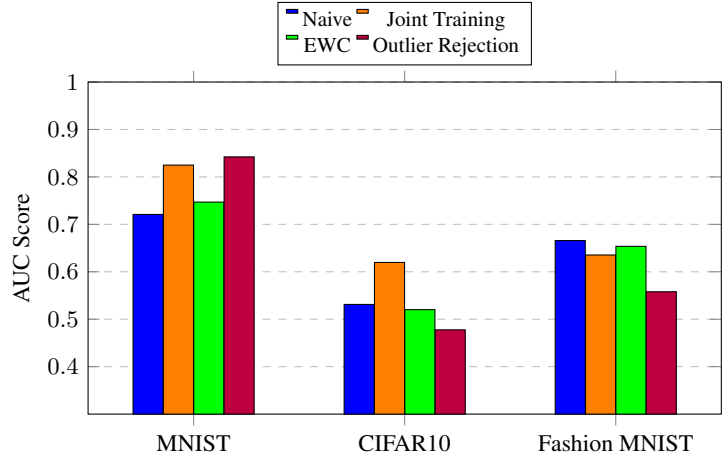
\begin{figure}[h]
    \centering
        \centering
        \begin{tikzpicture}
            \begin{axis}[
                height=6cm,
                width=10cm,
                ybar=0.1pt,
                bar width=0.4cm,
                enlarge x limits=0.25,
                symbolic x coords={MNIST, CIFAR10, Fashion MNIST},
                xtick=data,
                ylabel={AUC Score},
                ymajorgrids=true,
                grid style=dashed,
                ymin=0.3, 
                ymax=1, 
                ytick={0.4, 0.5, 0.6, 0.7, 0.8,0.9,1.0},
                legend image code/.code={
                \draw [#1] (0cm,0.1cm) rectangle (0.15cm,0.15cm); },
                x tick label style={rotate=0, anchor=north},
                legend style={at={(0.5,1.05)}, anchor=south, legend columns=2, nodes={scale=0.8, transform shape}},
                label style={font=\footnotesize},
                tick label style={font=\footnotesize}  
            ]
\addplot[ybar, fill=blue] coordinates {(MNIST,0.7209) (CIFAR10,0.5312) (Fashion MNIST,0.6658)};
\addplot[ybar, fill=orange] coordinates {(MNIST,0.8249) (CIFAR10,0.6197) (Fashion MNIST,0.6354)};
\addplot[ybar, fill=green] coordinates {(MNIST,0.7469) (CIFAR10,0.5202) (Fashion MNIST,0.6536)};
\addplot[ybar, fill=purple] coordinates {(MNIST,0.8422) (CIFAR10,0.4776) (Fashion MNIST,0.5579)};
            \addlegendimage{ybar,fill=blue}
            \addlegendentry{Naive}
            \addlegendimage{ybar,fill=orange}
            \addlegendentry{Joint Training}
            \addlegendimage{ybar,fill=green}
            \addlegendentry{EWC}
            \addlegendimage{ybar,fill=purple}
            \addlegendentry{Outlier Rejection}
            \end{axis}
        \end{tikzpicture}
        \caption{\textbf{AUC Scores (Exp 8)}}

    \label{experiment_8_appendix}
\end{figure}

\begin{figure}[h]
    \centering
    \begin{subfigure}{0.6\textwidth}
        \centering
        \begin{tikzpicture}
            \begin{axis}[
                height=6cm,
                width=10cm,
                ybar=0.1pt,
                bar width=0.4cm,
                enlarge x limits=0.25,
                symbolic x coords={MNIST, CIFAR10, Fashion MNIST},
                xtick=data,
                ylabel={AUC Score},
                ymajorgrids=true,
                grid style=dashed,
                ymin=0.3, 
                ymax=0.8, 
                ytick={0.4, 0.5, 0.6, 0.7, 0.8},
                legend image code/.code={
                \draw [#1] (0cm,0.1cm) rectangle (0.15cm,0.15cm); },
                x tick label style={rotate=0, anchor=north},
                legend style={at={(0.5,1.05)}, anchor=south, legend columns=2, nodes={scale=0.8, transform shape}},
                label style={font=\footnotesize},
                tick label style={font=\footnotesize}  
            ]
\addplot[ybar, fill=blue] coordinates {(MNIST,0.6350) (CIFAR10,0.5236) (Fashion MNIST,0.5280)};
\addplot[ybar, fill=orange] coordinates {(MNIST,0.5250) (CIFAR10,0.5358) (Fashion MNIST,0.3319)};
\addplot[ybar, fill=green] coordinates {(MNIST,0.5067) (CIFAR10,0.5828) (Fashion MNIST,0.5481)};
\addplot[ybar, fill=purple] coordinates {(MNIST,0.5913) (CIFAR10,0.5349) (Fashion MNIST,0.4541)};

            \addlegendimage{ybar,fill=blue}
            \addlegendentry{Naive}
            \addlegendimage{ybar,fill=orange}
            \addlegendentry{Joint Training}
            \addlegendimage{ybar,fill=green}
            \addlegendentry{EWC}
            \addlegendimage{ybar,fill=purple}
            \addlegendentry{Outlier Rejection}
            \end{axis}
        \end{tikzpicture}
        \caption{\textbf{AUC Scores (Exp 9)}}
    \end{subfigure}
    \caption{AUC Scores for Different Methods on MNIST, CIFAR10, and Fashion MNIST}
    \label{experiment_9_appendix}
\end{figure}

\begin{figure}[h]
    \centering
    \begin{subfigure}{0.6\textwidth}
        \centering
        \begin{tikzpicture}
            \begin{axis}[
                height=6cm,
                width=10cm,
                ybar=0.1pt,
                bar width=0.4cm,
                enlarge x limits=0.25,
                symbolic x coords={MNIST, CIFAR10, Fashion MNIST},
                xtick=data,
                ylabel={AUC Score},
                ymajorgrids=true,
                grid style=dashed,
                ymin=0.3, 
                ymax=0.8, 
                ytick={0.4, 0.5, 0.6, 0.7, 0.8},
                legend image code/.code={
                \draw [#1] (0cm,0.1cm) rectangle (0.15cm,0.15cm); },
                x tick label style={rotate=0, anchor=north},
                legend style={at={(0.5,1.05)}, anchor=south, legend columns=2, nodes={scale=0.8, transform shape}},
                label style={font=\footnotesize},
                tick label style={font=\footnotesize}  
            ]
\addplot[ybar, fill=blue] coordinates {(MNIST,0.6477) (CIFAR10,0.5473) (Fashion MNIST,0.5467)};
\addplot[ybar, fill=orange] coordinates {(MNIST,0.6519) (CIFAR10,0.5638) (Fashion MNIST,0.3892)};
\addplot[ybar, fill=green] coordinates {(MNIST,0.7081) (CIFAR10,0.5083) (Fashion MNIST,0.5632)};
\addplot[ybar, fill=purple] coordinates {(MNIST,0.7111) (CIFAR10,0.5472) (Fashion MNIST,0.6000)};

            \addlegendimage{ybar,fill=blue}
            \addlegendentry{Naive}
            \addlegendimage{ybar,fill=orange}
            \addlegendentry{Joint Training}
            \addlegendimage{ybar,fill=green}
            \addlegendentry{EWC}
            \addlegendimage{ybar,fill=purple}
            \addlegendentry{Outlier Rejection}
            \end{axis}
        \end{tikzpicture}
        \caption{\textbf{AUC Scores (Exp 10)}}
    \end{subfigure}
    \caption{AUC Scores for Different Methods on MNIST, CIFAR10, and Fashion MNIST}
    \label{experiment_10_appendix}
\end{figure}

\begin{figure}[h]
    \centering
    \begin{subfigure}{0.6\textwidth}
        \centering
        \begin{tikzpicture}
            \begin{axis}[
                height=6cm,
                width=10cm,
                ybar=0.1pt,
                bar width=0.4cm,
                enlarge x limits=0.25,
                symbolic x coords={MNIST, CIFAR10, Fashion MNIST},
                xtick=data,
                ylabel={AUC Score},
                ymajorgrids=true,
                grid style=dashed,
                ymin=0.3, 
                ymax=0.8, 
                ytick={0.4, 0.5, 0.6, 0.7, 0.8},
                legend image code/.code={
                \draw [#1] (0cm,0.1cm) rectangle (0.15cm,0.15cm); },
                x tick label style={rotate=0, anchor=north},
                legend style={at={(0.5,1.05)}, anchor=south, legend columns=2, nodes={scale=0.8, transform shape}},
                label style={font=\footnotesize},
                tick label style={font=\footnotesize}  
            ]
\addplot[ybar, fill=blue] coordinates {(MNIST,0.6116) (CIFAR10,0.5602) (Fashion MNIST,0.5582)};
\addplot[ybar, fill=orange] coordinates {(MNIST,0.5855) (CIFAR10,0.5499) (Fashion MNIST,0.3271)};
\addplot[ybar, fill=green] coordinates {(MNIST,0.5767) (CIFAR10,0.5601) (Fashion MNIST,0.4369)};
\addplot[ybar, fill=purple] coordinates {(MNIST,0.6568) (CIFAR10,0.6367) (Fashion MNIST,0.4555)};

            \addlegendimage{ybar,fill=blue}
            \addlegendentry{Naive}
            \addlegendimage{ybar,fill=orange}
            \addlegendentry{Joint Training}
            \addlegendimage{ybar,fill=green}
            \addlegendentry{EWC}
            \addlegendimage{ybar,fill=purple}
            \addlegendentry{Outlier Rejection}
            \end{axis}
        \end{tikzpicture}
        \caption{\textbf{AUC Scores (Exp 11)}}
    \end{subfigure}
    \caption{AUC Scores for Different Methods on MNIST, CIFAR10, and Fashion MNIST}
    \label{experiment_11_appendix}
\end{figure}

In Figure \ref{experiment_8_appendix}, the first thing of note is that \(\gamma\) is largely optimal at 0.2 in the case of MNIST and fashion MNIST. It is initially surprising that there is not a linear trend of \(\beta\) against AUC. However, it is likely that 0.2 is the tipping point between not enough labelled anomalies for the classifier to be able to properly distinguish normal from anomalous data and too many labelled anomalies that then violate the cluster assumption of semi-supervised learning \cite{ouali2020overview}.  

Additionally, for MNIST in particular, outlier rejection is very often the best continual learning method employed. However, when \(\gamma\) = 0.5, outlier rejection performs poorly across all datasets. This could be because of the overlap between labelled anomalies and labelled normal data in the latent space of the VAE with the addition of labelled anomalous data. According to Mathieu \textit{et al}. \cite{mathieu2019disentangling}, there needs to be a delicate balance of overlap between classes in a latent space for the VAE to create a meaningful representation of the inputs. Therefore, this overlap is not something that can be avoided without the latent space becoming something of a lookup table which then tends towards a replay buffer disguised as generative replay. 

\textbf{Ablation 4}

\begin{figure}[h]
    \centering
    \begin{subfigure}{0.6\textwidth}
        \centering
        \begin{tikzpicture}
            \begin{axis}[
                height=6cm,
                width=10cm,
                ybar=0.1pt,
                bar width=0.4cm,
                enlarge x limits=0.25,
                symbolic x coords={MNIST, CIFAR10, Fashion MNIST},
                xtick=data,
                ylabel={AUC Score},
                ymajorgrids=true,
                grid style=dashed,
                ymin=0.3, 
                ymax=0.8, 
                ytick={0.4, 0.5, 0.6, 0.7, 0.8},
                legend image code/.code={
                \draw [#1] (0cm,0.1cm) rectangle (0.15cm,0.15cm); },
                x tick label style={rotate=0, anchor=north},
                legend style={at={(0.5,1.05)}, anchor=south, legend columns=2, nodes={scale=0.8, transform shape}},
                label style={font=\footnotesize},
                tick label style={font=\footnotesize}  
            ]
\addplot[ybar, fill=blue] coordinates {(MNIST,0.6887) (CIFAR10,0.5251) (Fashion MNIST,0.5508)};
\addplot[ybar, fill=orange] coordinates {(MNIST,0.6953) (CIFAR10,0.5220) (Fashion MNIST,0.3972)};
\addplot[ybar, fill=green] coordinates {(MNIST,0.6789) (CIFAR10,0.5528) (Fashion MNIST,0.5152)};
\addplot[ybar, fill=purple] coordinates {(MNIST,0.7265) (CIFAR10,0.5329) (Fashion MNIST,0.5969)};

            \addlegendimage{ybar,fill=blue}
            \addlegendentry{Naive}
            \addlegendimage{ybar,fill=orange}
            \addlegendentry{Joint Training}
            \addlegendimage{ybar,fill=green}
            \addlegendentry{EWC}
            \addlegendimage{ybar,fill=purple}
            \addlegendentry{Outlier Rejection}
            \end{axis}
        \end{tikzpicture}
        \caption{\textbf{AUC Scores (Exp 12)}}
    \end{subfigure}
    \caption{AUC Scores for Different Methods on MNIST, CIFAR10, and Fashion MNIST}
    \label{experiment_12_appendix}
\end{figure}

\begin{figure}[h]
    \centering
    \begin{subfigure}{0.6\textwidth}
        \centering
        \begin{tikzpicture}
            \begin{axis}[
                height=6cm,
                width=10cm,
                ybar=0.1pt,
                bar width=0.4cm,
                enlarge x limits=0.25,
                symbolic x coords={MNIST, CIFAR10, Fashion MNIST},
                xtick=data,
                ylabel={AUC Score},
                ymajorgrids=true,
                grid style=dashed,
                ymin=0.3, 
                ymax=0.8, 
                ytick={0.4, 0.5, 0.6, 0.7, 0.8},
                legend image code/.code={
                \draw [#1] (0cm,0.1cm) rectangle (0.15cm,0.15cm); },
                x tick label style={rotate=0, anchor=north},
                legend style={at={(0.5,1.05)}, anchor=south, legend columns=2, nodes={scale=0.8, transform shape}},
                label style={font=\footnotesize},
                tick label style={font=\footnotesize}  
            ]
\addplot[ybar, fill=blue] coordinates {(MNIST,0.6352) (CIFAR10,0.6028) (Fashion MNIST,0.5609)};
\addplot[ybar, fill=orange] coordinates {(MNIST,0.6700) (CIFAR10,0.5538) (Fashion MNIST,0.3496)};
\addplot[ybar, fill=green] coordinates {(MNIST,0.6833) (CIFAR10,0.5910) (Fashion MNIST,0.5715)};
\addplot[ybar, fill=purple] coordinates {(MNIST,0.6817) (CIFAR10,0.5518) (Fashion MNIST,0.5546)};

            \addlegendimage{ybar,fill=blue}
            \addlegendentry{Naive}
            \addlegendimage{ybar,fill=orange}
            \addlegendentry{Joint Training}
            \addlegendimage{ybar,fill=green}
            \addlegendentry{EWC}
            \addlegendimage{ybar,fill=purple}
            \addlegendentry{Outlier Rejection}
            \end{axis}
        \end{tikzpicture}
        \caption{\textbf{AUC Scores (Exp 13)}}
    \end{subfigure}
    \caption{AUC Scores for Different Methods on MNIST, CIFAR10, and Fashion MNIST}
    \label{experiment_13_appendix}
\end{figure}

\begin{figure}[h]
    \centering
    \begin{subfigure}{0.6\textwidth}
        \centering
        \begin{tikzpicture}
            \begin{axis}[
                height=6cm,
                width=10cm,
                ybar=0.1pt,
                bar width=0.4cm,
                enlarge x limits=0.25,
                symbolic x coords={MNIST, CIFAR10, Fashion MNIST},
                xtick=data,
                ylabel={AUC Score},
                ymajorgrids=true,
                grid style=dashed,
                ymin=0.3, 
                ymax=0.8, 
                ytick={0.4, 0.5, 0.6, 0.7, 0.8},
                legend image code/.code={
                \draw [#1] (0cm,0.1cm) rectangle (0.15cm,0.15cm); },
                x tick label style={rotate=0, anchor=north},
                legend style={at={(0.5,1.05)}, anchor=south, legend columns=2, nodes={scale=0.8, transform shape}},
                label style={font=\footnotesize},
                tick label style={font=\footnotesize}  
            ]
\addplot[ybar, fill=blue] coordinates {(MNIST,0.6077) (CIFAR10,0.5377) (Fashion MNIST,0.5690)};
\addplot[ybar, fill=orange] coordinates {(MNIST,0.5812) (CIFAR10,0.5556) (Fashion MNIST,0.3841)};
\addplot[ybar, fill=green] coordinates {(MNIST,0.6266) (CIFAR10,0.5536) (Fashion MNIST,0.5646)};
\addplot[ybar, fill=purple] coordinates {(MNIST,0.6403) (CIFAR10,0.5983) (Fashion MNIST,0.4862)};

            \addlegendimage{ybar,fill=blue}
            \addlegendentry{Naive}
            \addlegendimage{ybar,fill=orange}
            \addlegendentry{Joint Training}
            \addlegendimage{ybar,fill=green}
            \addlegendentry{EWC}
            \addlegendimage{ybar,fill=purple}
            \addlegendentry{Outlier Rejection}
            \end{axis}
        \end{tikzpicture}
        \caption{\textbf{AUC Scores (Exp 14)}}
    \end{subfigure}
    \caption{AUC Scores for Different Methods on MNIST, CIFAR10, and Fashion MNIST}
    \label{experiment_14_appendix}
\end{figure}

\begin{figure}[h]
    \centering
    \begin{subfigure}{0.6\textwidth}
        \centering
        \begin{tikzpicture}
            \begin{axis}[
                height=6cm,
                width=10cm,
                ybar=0.1pt,
                bar width=0.4cm,
                enlarge x limits=0.25,
                symbolic x coords={MNIST, CIFAR10, Fashion MNIST},
                xtick=data,
                ylabel={AUC Score},
                ymajorgrids=true,
                grid style=dashed,
                ymin=0.1, 
                ymax=0.8, 
                ytick={0.1,0.2,0.3,0.4, 0.5, 0.6, 0.7, 0.8},
                legend image code/.code={
                \draw [#1] (0cm,0.1cm) rectangle (0.15cm,0.15cm); },
                x tick label style={rotate=0, anchor=north},
                legend style={at={(0.5,1.05)}, anchor=south, legend columns=2, nodes={scale=0.8, transform shape}},
                label style={font=\footnotesize},
                tick label style={font=\footnotesize}  
            ]
\addplot[ybar, fill=blue] coordinates {(MNIST,0.5432) (CIFAR10,0.4575) (Fashion MNIST,0.2932)};
\addplot[ybar, fill=orange] coordinates {(MNIST,0.5151) (CIFAR10,0.4399) (Fashion MNIST,0.2905)};
\addplot[ybar, fill=green] coordinates {(MNIST,0.5081) (CIFAR10,0.4394) (Fashion MNIST,0.2948)};
\addplot[ybar, fill=purple] coordinates {(MNIST,0.6145) (CIFAR10,0.4445) (Fashion MNIST,0.2993)};

            \addlegendimage{ybar,fill=blue}
            \addlegendentry{Naive}
            \addlegendimage{ybar,fill=orange}
            \addlegendentry{Joint Training}
            \addlegendimage{ybar,fill=green}
            \addlegendentry{EWC}
            \addlegendimage{ybar,fill=purple}
            \addlegendentry{Outlier Rejection}
            \end{axis}
        \end{tikzpicture}
        \caption{\textbf{AUC Scores (Exp 15)}}
    \end{subfigure}
    \caption{AUC Scores for Different Methods on MNIST, CIFAR10, and Fashion MNIST}
    \label{experiment_15_appendix}
\end{figure}

In this ablation study, the percentage of unlabelled anomalies in an experience \(\zeta\) is varied. Similarly to ablation study 3, this is varied for all experiences and not between experiences. As seen in Figure \ref{experiment_12_appendix}, as \(\zeta\) increases, AUC initially increases to a local maxima, and then decreases. The reason behind the decrease is likely due to violation of the cluster assumption of semi-supervised learning. This assumption postulates that similarly clustered inputs are contained within the same class \cite{ouali2020overview}. However, as more unlabelled anomalies are added, it is more likely that some are close to the decision boundary between normal and anomalous. Therefore adding these new points can move the decision boundary by expanding what the VAE reconstructs as anomalous. Since the VAE can then reconstruct some of the anomalous inputs, the ELBO for these then drops, meaning that they are more likely to be regarded as normal. 

However, the increase in performance that is observed only occurs up to a point, where model performance is highest when \(\zeta\) = 0.2. This is likely due to the unlabelled anomalies overlapping with the labelled anomalies leading to better separation between classes due to the limited amount of labelled anomalous data available (\(\gamma\) = 0.05 in these experiments). Once new anomalous data is introduced that doesn't fit into an existing labelled cluster, this can then lead to a decline in model performance. 

\section{Conclusion}
This work introduces and formalizes the novel problem of Continual Semi-Supervised Anomaly Detection (CSAD), a paradigm that integrates the complexities of semi-supervised learning, continual learning, and anomaly detection. The proposed approach, built upon a Variational Autoencoder (VAE) architecture with outlier rejection, demonstrates its efficacy in addressing the challenges of dynamic, real-world data streams. Key findings reveal the critical role of leveraging labelled data and effective anomaly handling, with our outlier rejection method often outperforming baseline methods such as Elastic Weight Consolidation (EWC) in several benchmark datasets.

Empirical results underscore the sensitivity of CSAD to varying labelled and unlabelled data distributions, highlighting the delicate balance required between labelled anomaly inclusion and the stability of latent space representations. While the proposed method shows promise, limitations such as reduced performance on high-dimensional datasets like CIFAR-10 point to the need for enhanced encoder-decoder architectures, such as convolutional layers for complex image data.

Future research should focus on refining the generative replay process, exploring advanced anomaly detection metrics, and incorporating more diverse datasets to validate broader applicability. Moreover, the extension of the framework to handle multi-modal data and adaptive hyperparameter tuning could significantly enhance its real-world usability. By laying the groundwork for CSAD, this study paves the way for robust anomaly detection solutions in dynamic environments, aligning closer to real-world operational constraints.

\bibliography{Bibliography}

\end{document}